# Human Gait Recognition Using Bag of Words Feature Representation Method

Nasrin Bayat [1], Elham Rastegari [2], Qifeng Li[3]

[1] Department of Electrical and Computer Engineering

University of Central Florida

Orlando, FL 32816, USA

[2] Department of Business, Intelligence and Analytics

Creighton University

Omaha, NE 68178, USA

[3] Department of Electrical and Computer Engineering

University of Central Florida

Orlando, FL 32816, USA

# ABSTRACT

In this paper, we propose a novel gait recognition method based on a bag-of-words feature representation method. The algorithm is trained, tested and evaluated on a unique human gait data consisting of 93 individuals who walked with comfortable pace between two end points during two different sessions. To evaluate the effectiveness of the proposed model, the results are compared with the outputs of the classification using extracted features. As it is presented, the proposed method results in significant improvement accuracy compared to using common statistical features, in all the used classifiers.

# INTRODUCTION

Studies show individuals can be recognized based on their gait (Nemes et al. 2020). Using mobile devices to identify different individuals is a critical and challenging research subject with several applications in biometric authentication (Al et al. 2017), personalized interfaces, smart surroundings, and healthcare (Zaki et al. 2020), (Bayat et al. 2017). A human gait recognition system requires four phases: data collection, feature representation, dimension reduction, and classification, regardless of the underlying techniques used. There are three primary types of approaches to collect human gait data in the literature: image processing-based, floor sensors-based, wearable sensors-based. Image processing systems use one or more optic sensors to collect data on the subject's gait and use digital image processing to take objective measurements of the various parameters. The floor sensors-based systems rely on sensors placed along the floor, where gait data is collected using pressure sensors and ground reaction force sensors, which detect the force exerted by the subject's feet on the floor while walking (Muro et al. 2014). Wearable sensors-based methods, capture pure walking dynamics using sensors directly worn by the user, with no need to equip the environment, allowing for ubiquitous recognition (Marsico and Mecca, 2019). In this study we use wearable sensors-based method. A smart phone connected to the right thigh of each participant of this study, collected the accelerometer data.

Originally, the bag of words method was used to index text materials. The bag of words method has also been effectively utilized in computer vision for over twenty years, and it is also known as the Bag of Features or Bag of Visual Words in the literature on the subject (Gabryel, 2018). In addition, bag of words variable representation is used for physical activity identification (Kheirkhahan et al. 2017) and the results outperform traditional wrist variables by 8% and 5%, respectively, for sedentary and locomotor physical activity detection. Based on the effectiveness of this approach in the other domains, we are applying it to the human gait recognition domain.

For accelerometer data, there are two basic feature representation methods: gait-cycle-based and frame-based methods. In the frame-based methods, raw signals are divided into several frames or windows using statistical features. A feature vector is extracted for each frame. Finally, the feature vector sequence gives information on the gait, which will be used for classification (Wan et al. 2018). In this paper, we propose a new method for feature representation of accelerometer data (see Figure 1). We first get raw data from the accelerometer, then preprocess it to extract features. In the next step, we convert 15 features to words. Finally, we use different classification methods and compare the results with the statistical features method.

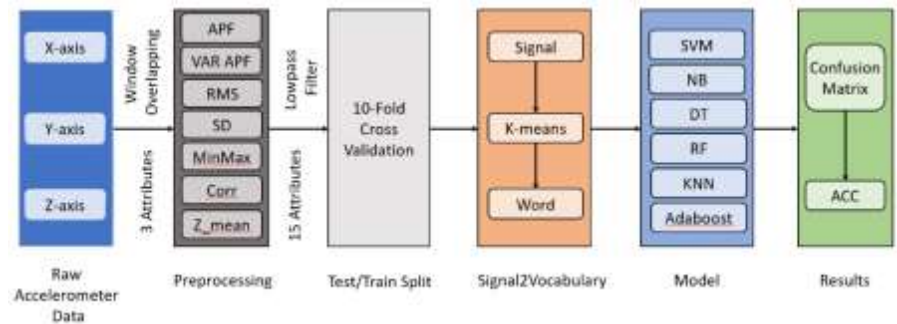

Figure 1. General pipeline of the proposed human gait recognition method

The major contributions of this work are summarized as follows:
We apply bag of words feature representation to human gait recognition using smart phone's accelerometer data.
We use the bag-of-words feature representation method to construct a human gait recognition model that surpasses the common statistical features method with up to 24 ±7.9 percent more accuracy.
The rest of the paper is organized as follows: In section 3, we cover the data collection methods. Data preprocessing is discussed in section 4. In section 5, we describe out feature extraction strategy. Then, our proposed method for feature representation is presented in section 6. Finally, simulation and conclusion are discussed in section 7 and 8, respectively.

## Data Collection

In this paper, we use the human gait database for normal walk (Vajdi et al. 2020) that was collected in our previous study. The data for the study is available for public upon request (Bayat, 2019). The 93 people participated in this experiment and walked at a comfortable speed in two different sessions. Each session was conducted on a different day. Two distinct smart phones (iPhone 6S) were used to record the data in each session, one was placed on the left waist and the other on the right thigh (see Figure 2). In terms of gender, the dataset is evenly distributed, with a male-to-female ratio close to one. Research studies have shown men and women have different walking patterns (Cho et al. 2004). Thus, the number of men and women should be almost the same to have a balanced dataset. Each participant walked 320 meters forward and backward between two endpoints. The experiment took place in the same site for all of the subjects, at a sea level of 0. We utilized the SensorLog application (1.9.7 version), which was built and tested for the IOS framework, to capture the data. A single log file was connected with each subject. Every log file had 3 features that were recorded and calculated internally by the SensorLog app. The features include acceleration along the x, y, and z axes, in which, the user's horizontal/sideways movement, upward/downward movement, and forward/backward movement were captured on x,y and z axes respectively. A sample of data related to one user is illustrated (see Figure 3). The data sampling rate is 100 samples per second.

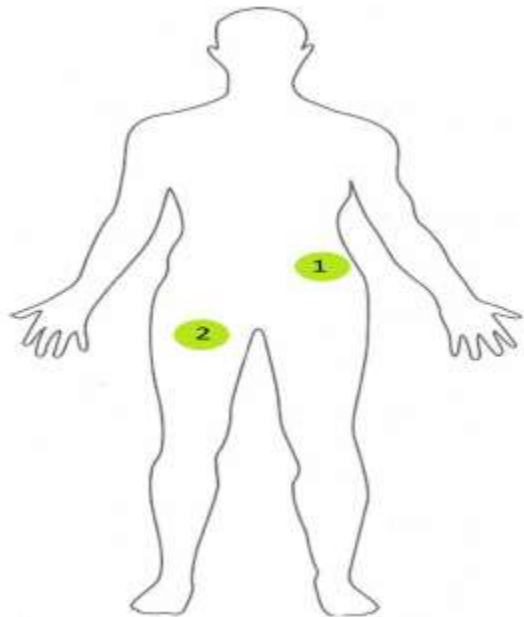

Figure 2. One of the smartphones was connected to the point 1 (left waist) and the other one was connected to the point 2 (right thigh).

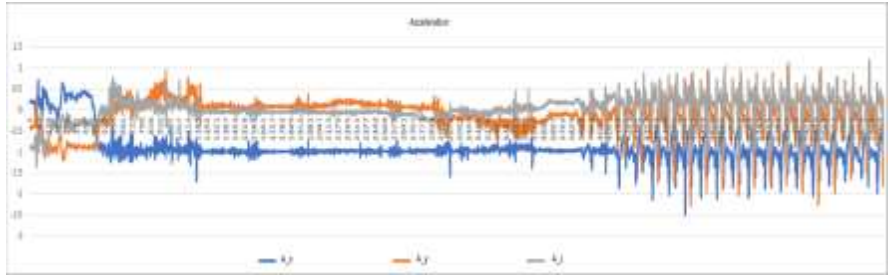

Figure 3. A user's accelerometer data along x, y, and z axes over time. The horizontal axis shows time in milliseconds. The vertical axis shows acceleration.

## Data Preprocessing

We only use the data of the right thigh of 42 people from first session since some of the participants did not attend the second session, and the data of the left waist was not complete for all the participants. The initial dataset contains time-series records from accelerometer sensors in three dimensions: X-axis, Y-axis, and Z-axis. Linear acceleration attributable to body motion and gravity are combined in each time series. In order to separate the high frequency components associated with body motion from the low frequency gravity component in each time series, we design and use a digital low pass filter (Bayat et al. 2014). The high frequency component, or AC component, is usually related to the subject's dynamic movements. The low-frequency component, or DC component, is primarily affected by gravity (Gyllensten and Bonomi, 2011). We define our low-pass filter in Eq.1:

$$D_{DC}[m] = M_1 D[m] + N_1 D_{DC}[m-1]$$

*(1)*

Where D is the raw data and $D_{DC}$ is the low pass filtered data. The sampling rate and cut-off frequency determine the value of the filter coefficients $M_1$ and $N_1$. The best cutoff frequency for excluding the gravity component alone would be between 0.1 and 0.5 HZ (Fujiki, 2010). Then, we came up at the following solution for the low-pass filter coefficients with a cutoff frequency of 0.25 Hz and a sampling rate of 100 samples per second: $N_1 = 0.9843$, $M_1 = 0.0158$. We subtract the low pass filtered data from the original signal in three axes to find the body acceleration components. On each time series, low pass and high pass filtering create gravity and body acceleration data in separate directions. The efficiency of the low pass filter in isolating the gravity and body acceleration is provided (see Figure 4). After applying the low-pass filter, two more time series $D_{DCi}$ and $D_{ACi}$ are produced for each time series $D_i$, representing the DC and AC component along each axis, respectively. To recognize people based on their walking pattern, we just need the AC component, and the magnitude of acceleration in all directions as investigated in our previous study.

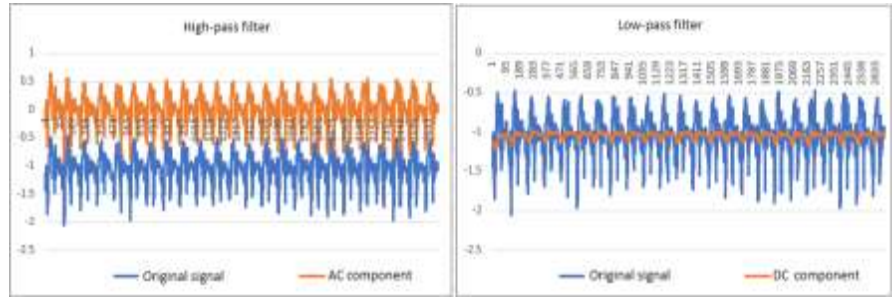

Figure 4. Effect of high pass filter (left figure) and low pass filter (right figure) on the accelerometer data along x axis. A signal containing only the body acceleration is produced by a high pass filter, while a signal having only the gravitational acceleration is produced by a low pass filter.

## Feature Extraction

Feature extraction is crucial in machine learning; More discriminative and descriptive features could lead to better prediction results. In order to extract features, first we pre-process the data as discussed in the previous section, then we apply a window overlapping technique to capture local changes that might be smeared if we look at the signal as a whole. It means, we compute features from a series of windows of a specific size by partitioning the dataset into smaller subgroups and then window them independently in the window overlapping method. For feature extraction from each of the time series, we chose a window of 100 samples, which corresponds to 1 second of accelerometer data. Features are generated with 50 percent overlap between subsequent windows to reduce information loss at each window margins.

One of the features we employ is the average number of peaks that occur in each window (Bayat et al. 2014). This attribute is known as the average of peak frequency (APF). Another feature that can assess how far a set of APF along the x-axis, y-axis, and z-axis is spread out is the variance of APF along three axes (VarAPF). The gravity component can be estimated using the average on each axis during a certain time period. As a result, one of the features is the average value of each window, which we compute it along z axis ($Z_m$). To derive two other features, the root mean square (RMS) value and the standard deviation (SD) are computed for each window. In addition, two more features are employed to improve human gait recognition: the difference between each window's maximum and minimum (Minmax), and the correlation between distinct axes (Corr). This way, we get a 15-dimensional feature vector (see Table 1).

Table 1: Description and Number of The Extracted Features

| Feature | Description | No. of features |
|---|---|---|
| APF | Average number of peaks | 3 |
| VarAPF | Variance of APF | 1 |
| $Z_m$ | Average on z axis | 1 |
| RMS | Root mean square | 3 |
| SD | Standard deviation | 3 |
| Minmax | The min-max for each window | 3 |
| Corr | Correlation between y and z axes | 1 |

At the next step, each subject's data is divided to 10 folds. The 10-fold cross validation is used for splitting data in train and test sets. In this way, the first 9 folds of all the subjects are concatenated to each other, and the resulting array is used as a train set. The remaining 10 percent of the data of all subjects is used as a test data.

# BAG OF WORDS FEATURE REPRESENTATION

In this section, the acceleration data recorded by smartphone sensors is transformed to a bag of words per individual and used to classify 42 people walking's patterns. The bag-of-words feature representation methodology can be explained in three steps (Rastegari and Ali, 2020): First, every observation is viewed as a sequence of time series data. Then, using a sliding window, each time series sequence is separated into overlapping windows, and comparable subsequences are clustered together and the same term is assigned to each cluster's subsequences. The first step is done in the feature extraction part. The next step is to come up with a vocabulary. We assume, each individual's walking signal is made up of multiple subsequences. And each subsequence is considered as a word. We combined all of the subsequences from all of the subjects and grouped subsequences based on their similarity. The K-means clustering algorithm is used to obtain k patterns for signal subsequences and generate the vocabulary.

The number of clusters (k) used in the above-mentioned clustering methods is crucial and can have an impact on the final result. Choosing the right number of clusters can serve a number of objectives, including obviating the necessity for any feature selection or feature reduction techniques. It also aids us in dealing with noise and over-fitting problems. Choosing a large number of clusters will result in over-fitting and a model that is not noise resistant. As a result, in the presence and absence of noise, the same subsequence will be assigned to various clusters. Whereas, picking a small number of clusters leads in a low distinguishing power. The elbow method (Rastegari and Ali, 2020) is used to find the optimal value for the number of clusters (k). The number of words required to depict all individuals' walking patterns is determined by the optimal number of clusters. We use the elbow method with the Within Cluster Sum of Square (WCSS) criterion as explained below.

To demonstrate how the WCSS criterion works, assume there are M clusters, each cluster has a center point $C_j$. Each subsequence falls into one of M clusters, We calculate the sum of squares of each subsequence's distance from its center point $C_j$ in the $j_{th}$ cluster. The sum of squares for the $j_{th}$ cluster $CSS_j$, is as follows:

$$CSSj = \sum_{i=0}^{N} \sqrt{\sum_{j=0}^{M}(Cj - nji)^2}$$

*(2)*

To calculate WCSS, the sum of square values for all clusters are added together and averaged over the number of clusters. We can see that as the number of clusters increase, the value of WCSS decreases. The optimal number of clusters is where WCSS value does not decrease significantly. A number between k = 10 and k = 20 is likely to be the optimal number of clusters (see Figure 5). We did the classification with all the values in this range. Based on the results k = 10 is the best number of clusters.

After the clustering job is completed, each cluster centroid is given a word, resulting in a vocabulary of k words ($v_0$, $v_1$, ..., $v_k$).

As a final step, we assign a word to each subsequence's cluster centroid. Then, every person's walking pattern is converted into a collection of words. In this way, for each person, we create a k-element vector M.

## Modeling

The objective of our model is to recognize people from each other based on their walking patterns. We assess the performance of the multiple classifiers on the generated features. The following classifiers are selected based on their effectiveness in gait classification problem: Adaboost, K Nearest Neighbors (KNN), Random Forest, Decision Tree, Naive Bayes, Linear and rbf Support Vector Machine (SVM). We use accuracy to evaluate the performance of the proposed features.

The accuracy refers to the percentage of correct predictions made by our model (see Table 2).

To evaluate the performance of the bag of words feature representation, we compare the classification performance using these proposed features with the results of classification obtained using statistical features (see Table 3).

The accuracy of classification using statistical features for all the employed classifiers on average is lower than our proposed method. In addition, we provide confusion matrix of Naive Bayes classifier, for both methods (see Figures 6 and 7). We can see a lot of misclassifications in the confusion matrix of classification using statistical features. Additionally, the accuracy of classification using statistical features is 24±7.9 percent lower than our proposed method.

## Conclusion

In this paper, we developed a bag-of-words based features extraction method for human gait recognition problems. we used a dataset collected using built-in smartphone accelerometers during normal walking from 42 healthy subjects. Each individual's walking time series was then converted into signal subsequences using an overlapping sliding window. Then, each subsequence was characterized
using few statistical descriptors and similar subsequences were assigned the same word. Therefore, each person's data is
converted into a bag of words. We evaluated the bag-of-words method for automatic identification of subjects
using accelerometers data collected from right thighs. The results of this preliminary experiment indicate that the classification based on bag-of-words feature representation method outperforms classification method based on the statistical features. The results also suggest the Random Forest, and SVM classifiers had the best performance using both methods.

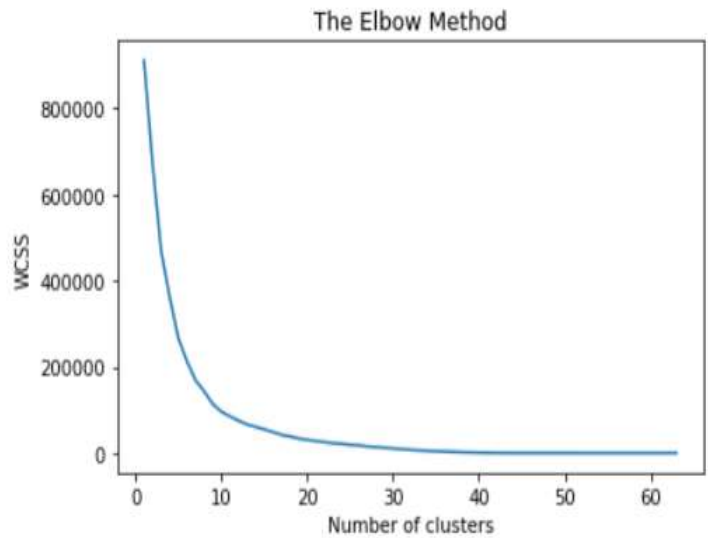

Figure 5. WCSS-based elbow method

Table 2: Accuracy of Classification Using Bag of Words Feature Representation.

| Classifier | Accuracy |
| --- | --- |
| LinearSVM | 0.976 |
| rbfSVM | 0.922 |
| NaiveBayes | 0.998 |
| DecisionTree | 0.992 |
| RandomForest | 0.999 |
| KNN | 0.993 |
| Adaboost | 0.761 |

Table 3: Accuracy of Classification Using Statistical Features.

| Classifier | Accuracy |
|---|---|
| LinearSVM | 0.781 |
| rbfSVM | 0.806 |
| NaiveBayes | 0.613 |
| DecisionTree | 0.686 |
| RandomForest | 0.750 |
| KNN | 0.772 |
| Adaboost | 0.549 |

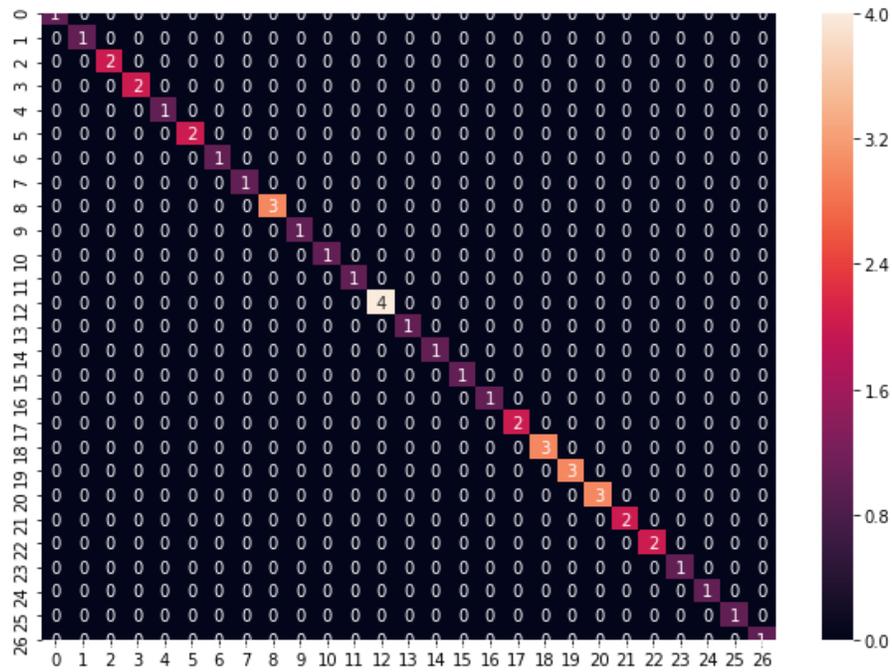

Figure 6. Confusion matrix of Naive Bayes algorithm, using Bag of Words Feature Representation.

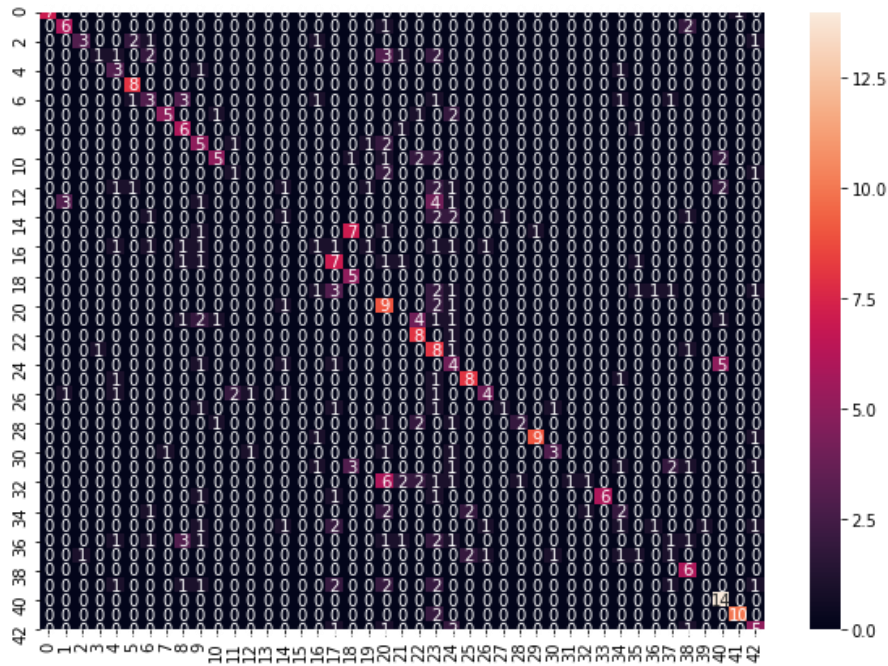

Figure 7. Confusion matrix of Naive Bayes algorithm, using statistical features and k=10

# REFERENCES

Al Kork, S.K., Gowthami, I., Savatier, X., Beyrouthy, T., Korbane, J.A. and Roshdi, S. (2017). Biometric database for human gait recognition using wearable sensors and a smartphone. [online] IEEE Xplore. Available at: https://ieeexplore.ieee.org/document/8095329 [Accessed 3 Feb. 2022].

Bayat, A. (2021). Human_Gait_Data. [online] GitHub. Available at: https://github.com/Bayat-Ak/Human_Gait_Data [Accessed 4 Feb. 2022].

Bayat, A., Hossein Bayat, A. and Ghasemi, A. (2017). Classifying Human Walking Patterns using Accelerometer Data from Smartphone. International Journal of Computer Science and Mobile Computing, [online] 6, pp.78–83. Available at: https://ijcsmc.com/docs/papers/December2017/V6I12201714.pdf [Accessed 4 Feb. 2022].

Bayat, A., Pomplun, M. and Tran, D.A. (2014). A Study on Human Activity Recognition Using Accelerometer Data from Smartphones. Procedia Computer Science, 34, pp.450–457.

Cho, S.H., Park, J.M. and Kwon, O.Y. (2004). Gender differences in three dimensional gait analysis data from 98 healthy Korean adults. Clinical Biomechanics (Bristol, Avon), [online] 19(2), pp.145–152. Available at: https://www.ncbi.nlm.nih.gov/pubmed/14967577.

Fujiki, Y. (2010). iPhone as a Physical Activity Measurement Platform CHI 2010: Student Research Competition (Spotlight on Posters Days 1 & 2). [online]

Available at: http://www.cpl.uh.edu/publication_files/C58B.pdf [Accessed 4 Feb. 2022].

Gabryel, M. (2018). The Bag-of-Words Method with Different Types of Image Features and Dictionary Analysis. J. Univers. Comput. Sci. [online] Available at: https://www.semanticscholar.org/paper/The-Bag-of-Words-Method-with-Different-Types-of-and-Gabryel/4edcc68bac9743b49cbca17a8f2d50596d754a2e [Accessed 4 Feb. 2022].

Gyllensten, I.C. and Bonomi, A. (2011). Identifying Types of Physical Activity With a Single Accelerometer: Evaluating Laboratory-trained Algorithms in Daily Life. IEEE Transactions on Biomedical Engineering. [online] Available at: https://www.semanticscholar.org/paper/Identifying-Types-of-Physical-Activity-With-a-in-Gyllensten-Bonomi/340f062214878596baf4d297d5ec723d57a8cc65 [Accessed 4 Feb. 2022].

Johns Hopkins University. (n.d.). A bag-of-words approach for assessing activities of daily living using wrist accelerometer data. [online] Available at: https://jhu.pure.elsevier.com/en/publications/a-bag-of-words-approach-for-assessing-activities-of-daily-living-/fingerprints/ [Accessed 4 Feb. 2022].

Marsico, M.D. and Mecca, A. (2019). A Survey on Gait Recognition via Wearable Sensors. ACM Computing Surveys, 52(4), pp.1–39.

Muro-de-la-Herran, A., Garcia-Zapirain, B. and Mendez-Zorrilla, A. (2014). Gait Analysis Methods: An Overview of Wearable and Non-Wearable Systems, Highlighting Clinical Applications. Sensors, [online] 14(2), pp.3362–3394. Available at: https://www.ncbi.nlm.nih.gov/pmc/articles/PMC3958266/.

Nemes, S. and Antal, M. (2020). Feature Learning for Accelerometer based Gait Recognition. arXiv:2007.15958 [cs]. [online] Available at: https://arxiv.org/abs/2007.15958 [Accessed 3 Feb. 2022].

Rastegari, E. and Ali, H. (2020). A bag-of-words feature engineering approach for assessing health conditions using accelerometer data. Smart Health, 16, p.100116.

Vajdi, A., Zaghian, M.R., Farahmand, S., Rastegar, E., Maroofi, K., Jia, S., Pomplun, M., Haspel, N. and Bayat, A. (2020). Human Gait Database for Normal Walk Collected by Smart Phone Accelerometer. arXiv:1905.03109 [cs, eess]. [online] Available at: https://arxiv.org/abs/1905.03109 [Accessed 4 Feb. 2022].

Wan, C., Wang, L. and Phoha, V.V. eds., (2019). A Survey on Gait Recognition. ACM Computing Surveys, 51(5), pp.1–35.

Zaki, T.H.M., Sahrim, M., Jamaludin, J., Balakrishnan, S., Asbulah, L.H. and Hussin, F.S. (2020). The Study of Drunken Abnormal Human Gait Recognition using Accelerometer and Gyroscope Sensors in Mobile Application. 2020 16th IEEE International Colloquium on Signal Processing & Its Applications (CSPA). [online] Available at: https://www.semanticscholar.org/paper/The-Study-of-Drunken-Abnormal-Human-Gait-using-and-Zaki-Sahrim/fd6c478d7db79dbcc2cdabba8d512f05acc1dbd7 [Accessed 4 Feb. 2022].